\documentclass[conference]{IEEEtran}
\ifCLASSINFOpdf
\else
\fi
\usepackage{color}
\usepackage{algorithm,algorithmicx}
\usepackage{algpseudocode,empheq,hhline}
\usepackage{amsmath,amsthm,amssymb,url,graphicx,rotating,ifthen,epsfig,array,color}
\usepackage{float}
\usepackage{dsfont}
\usepackage{mathtools}
\usepackage{multirow}
\usepackage{mathtools}
\usepackage{soul}
\usepackage{calligra}
\usepackage{cite}
\usepackage{colortbl}
\usepackage{rotating,longtable}
\usepackage{tabulary}
\usepackage{balance}
\usepackage{todonotes}

\def\longversion{0} 

\DeclareMathOperator*{\argmin}{arg\,min}
\DeclareMathOperator*{\argmax}{arg\,max}
\def\N{{\mathbb N}}

\begin{document}
%
\title{Rolling Horizon Coevolutionary Planning for Two-Player Video Games}


\author{\IEEEauthorblockN{Jialin Liu}
\IEEEauthorblockA{University of Essex\\
Colchester CO4 3SQ\\
United Kingdom\\
jialin.liu@essex.ac.uk}
\and
\IEEEauthorblockN{Diego P\'erez-Li\'ebana}
\IEEEauthorblockA{University of Essex\\
Colchester CO4 3SQ\\
United Kingdom\\
dperez@essex.ac.uk}
\and
\IEEEauthorblockN{Simon M. Lucas}
\IEEEauthorblockA{University of Essex\\
Colchester CO4 3SQ\\
United Kingdom\\
sml@essex.ac.uk}
}

%


\maketitle

\begin{abstract}
This paper describes a new algorithm for 
decision making in two-player real-time video games.  As with Monte Carlo Tree Search, the algorithm can be used without heuristics and has been developed for use in general video game AI.

The approach is to extend recent work on rolling horizon evolutionary planning, which has been shown to work well for single-player
games, to two (or in principle many) player games.  To select an action the algorithm co-evolves
two (or in the general case $N$) populations, one for each player, where each individual is a sequence of actions for the
respective player.  The fitness of each individual is evaluated by playing it against
a selection of action-sequences from the opposing population.  When choosing an action
to take in the game, the first action is chosen from the fittest member of the population for
that player.

The new algorithm is compared with a number of general video game AI
algorithms on three variations of a two-player space battle game,
with promising results.

\end{abstract}


%
\IEEEpeerreviewmaketitle


\section{Introduction}\label{sec:intro}
Since the dawn of computing, games have provided an excellent test bed for 
AI algorithms, and increasingly games have also been a major AI application area.
Over the last decade progress in game AI has been significant, with Monte Carlo Tree Search dominating many areas since 2006\cite{coulom2006efficient,kocsis2006improved,gelly2006modification,chaslot2006monte,chaslot2008monte}, and more recently deep reinforcement learning has provided amazing results, both in stand-alone mode in video games, and in combination with MCTS in the shape of AlphaGo\cite{Silver}, which achieved one of the greatest steps forward in a mature and competitive area of AI ever seen. AlphaGo not only beat Lee Sedol,
previous world Go champion, but also soundly defeated all leading Computer Go bots, many of which had been in development for more than a decade.
Now AlphaGo is listed at the second position among current strongest human players\footnote{http://www.goratings.org}.

With all this recent progress a distant observer could be forgiven for thinking that
Game AI was starting to become a solved problem, but with smart AI as a baseline the challenges become
more interesting, with ever greater opportunities to develop games that depend on AI either at the design
stage or to provide compelling gameplay.

The problem addressed in this paper is the design of general algorithms for
two-player video games. As a possible solution, we introduce a new algorithm:
Rolling Horizon Coevolution Algorithm (RHCA). This only works for cases when
the forward model of the game is known
and can be used to conduct what-if simulations (roll-outs) much faster
than real time. In terms of the General Video Game AI (GVGAI) competition\footnote{http://gvgai.net}
series, this is known as the two-player planning track\cite{perez20152014}.
However, this track was not available at the time of writing this paper,
so we were faced with the choice of using an existing two-player video game,
or developing our own.  We chose to develop our own set simple battle game, 
bearing some similarity to the original 1962 Spacewar game\footnote{https://en.wikipedia.org/wiki/Spacewar\_(video\_game)} though
lacking the fuel limit and the central death star with the gravity field.  This approach
provides direct control over all aspects of the game and enables us to
optimise it for fast simulations, and also to experiment with parameters to
test the generality of the results.

The two-player planning track is interesting because it offers
the possibility of AI which can be instantly smart with no prior training 
on a game, unlike Deep Q Learning (DQN) approaches \cite{mnih2015human} which require extensive training before
good performance can be achieved.  Hence the bots developed using our planning
methods can be used to provide instant feedback to game designers on aspects
of gameplay such as variability, challenge and skill depth.

The most obvious choice for this type of AI is Monte Carlo Tree Search (MCTS),
but recent results have shown that for single-player video games,
Rolling Horizon Evolutionary Algorithms (RHEA) are competitive.  Rolling horizon evolution works 
by evolving a population of action sequences, where the length of each sequence
equals the depth of simulation.  This is in contrast to MCTS where by default 
(and for reasons of efficiency, and of having enough visits to a node to make the statistics informative) 
the depth of tree is usually shallower than 
the total depth of the rollout (from root to the final evaluated state).

The use of action-sequences rather than trees as the core unit of evaluation has strengths and weaknesses.  Strengths include simplicity and efficiency, but the main weakness is that the individual sequence-based approach would by default ignore the savings possible by recognising shared prefixes, though prefix-trees can be constructed for evaluation purposes if it is more efficient to do so\cite{perez2015open}.
A further disadvantage is that the system is not directly utilising tree statistics to make more informed decisions, though given the limited simulation budget typically available in a real-time video game, the value of these statistics may be outweighed by the benefits of the rolling horizon approach.

Given the success of RHEA on single-player games, the question naturally
arises of how to extend the algorithm to two-player games (or in the general
case $N$-player games), with the follow-up question of how well such
an extension works.  The main contributions of this paper are the algorithm, RHCA, and the results of initial tests on
our two-player battle game.  The results are promising, and suggest that RHCA
could be a natural addition to a game AI practitioner's toolbox.

The rest of this paper is structured as follows: Section \ref{sec:back} provides
more background to the research, Section~\ref{sec:battle} describes the battle
games used for the experiments, Section~\ref{sec:cont} describes the controllers used in the experiments, Section~\ref{sec:xp} presents the results and Section~\ref{sec:conc} concludes.

\def\oldthings{
\section{OLD THINGS TO BE MOVED OR DELETED}
The past work in Game Artificial Intelligence is mostly around two-player board games or single-player games, including single-player video games or some classic problems, such as Traveling Salesman Problem.

Monte Carlo Tree Search (MCTS) is a powerful tree search method for finding optimal decisions by randomly sampling in the given decision space.
MCTS is widely used in many difficult games and outperforms alpha-beta when the evaluation function is hard to obtain/design or expensive to compute.
Gelly et al.\cite{gelly2007combining,lee2009computational} applied Upper Confidence Bound for Trees (UCT) to computer Go and resulted in the program \emph{MoGo}, which achieved the first win in 9x9 against a top profession human player at the game of Go, without handicap, and with large handicap in 19x19.
Recently, \emph{AlphaGo}\cite{Silver} realised the first win in full board against a top profession human player, without handicap by combining several learning tools, including imitation learning on deep Convolution Networks, learning by self-play, value function learning and MCTS.

By combining deep Convolution Network to Reinforcement Learning (RL), Deep Q-Network (DQN)\cite{Mnih2015} reached human-level performance on a variety of Atari games using limited prior knowledge (pixels and game scores).
Recently, \cite{Lillicrap2016} adapted DQN to the continuous action domain and achieved a stable performance on diverse simulated physics problems.
Both work \cite{Mnih2015,Lillicrap2016} solved different tasks using a single learning algorithm, which is the main task of General Game Playing.

The first General Game Playing competition, which mainly about two-player board game, is organized in Association for the Advancement of Artificial Intelligence (AAAI) in 2005\cite{genesereth2005general}.
Perez et al. carried out the General Video Game Playing (GVGP) framework and organized the first General Video Game AI (GVG-AI) Competition in 2014\cite{perez20152014}.
In this work, we introduced two Rolling Horizons Evolution Algorithms and compared them to one of the best algorithms in the GVG-AI Competition on a two-player battle game and its variants.

The paper is structured as follows.
This section reviews the current state of the art in Game AI. 
Section \ref{sec:back} present the concepts of algorithms used in this work. 
This includes applying Open Loop control on MCTS, Coevolutionary Generic Algorithm, and Rolling Horizon Planning.
Section \ref{sec:battle} describes the framework.
Then, Section \ref{sec:cont} explains the algorithms used in the experiments.
Experimental results are presented and discussed in Section \ref{sec:xp}.
Finally, Section \ref{sec:conc} concludes the work and discusses the possible future work.
}

\section{Background}\label{sec:back}

\subsection{Open Loop control}
Open loop control refers to those techniques which action decision mechanism is based on executing sequences of actions determined independently from the states visited during those sequences. Weber discusses the difference between closed and open loop in~\cite{weber2010optimization}. An open loop technique, Open Loop Expectimax Tree Search (OLETS), developed by Cou\"{e}toux, won the first GVGAI competition in 2014~\cite{perez20152014}, an algorithm inspired by Hierarchical Open Loop Optimistic Planning (HOLOP,~\cite{ICAPS124697}). Also in GVGAI, Perez et al.~\cite{perez2015open} discussed three different open loop techniques, then trained them on 28 games of the framework, and tested on 10 games. 

The classic closed loop tree search is efficient when the game is deterministic, i.e., given a state $s$, $\forall a \in A(s)$ ($A(s)$ is the set of legal actions in $s$), the next state $s' \gets S(a,s)$ is unique. In the stochastic case, given a state $s$, $\forall a \in A(s)$ ($A(s)$ is the set of legal actions in $s$), the state $s' \gets S(a,s)$ is drawn with a probability distribution from the set of possible future states.

\def\remove{
\cite{lucas2014fast} introduced an adaptive MCTS using $(1+1)$-ES to optimise the control parameters at each iteration to improve its performance.
The adaptive MCTS significantly outperformed the standard MCTS on the classic Mountain Car benchmark and a simplified version of Space Invaders.}

\subsection{Rolling Horizon planning}
Rolling Horizon planning, sometimes called Receding Horizon Control or Model Predictive Control (MPC), is commonly used in industries for making decisions in a dynamic stochastic environment. 
In a state $s$, an optimal input trajectory is made based on a forecast of the next $t_h$ states, where $t_h$ is the tactical horizon.
Only the first input of the trajectory is applied to the problem.
This procedure repeats periodically with updated observations and forecasts.

A rolling horizon version of an Evolutionary Algorithm that handles macro-actions was applied to the Physical Traveling Salesman Problem (PTSP) as introduced in~\cite{perez2013rolling}.
Then, RHEA was firstly applied on general video game playing by Perez et al. in \cite{perez2015open} and was shown to be the most efficient evolutionary technique on the GVGAI Competition framework.

\section{Battle games}\label{sec:battle}

Our two-player space battle game could be viewed as derived from the original Spacewar (as
mentioned above).  We then 
experimented with a number of variations to bring out some strengths and weaknesses
of the various algorithms under test.  A key finding is that the rankings of the algorithms depend
very much on the details of the game; had we just tested on a single version
of the game the conclusions would have been less robust and may have shown RHCA 
in a false light.
The agents are given full information about the game state and make their actions
simultaneously: the games are symmetric with perfect and incomplete information.
Each game commences with the agents in random symmetric positions to provide 
varied conditions while maintaining fairness.

\ifthenelse{\longversion=1}{
\subsection{A simple battle game without weapons}\label{sec:game1}
}{
\subsection{A simple battle game}\label{sec:game1}
}
First, a simple version without missiles is designed, referred as $G_1$.
Each spaceship, either owned by the first (green) or second (blue) player has the following
properties:
\begin{itemize}
\item has a maximal speed equals to $3$ units distance per game tick;
\item slows down over time;
\item can make a clockwise or anticlockwise rotation, or to thrust at each game tick.
\end{itemize}
Thus, the agents are in a fair situation.

\subsubsection{End condition}\label{sec:end1}
\begin{figure}[h]
\centering
\caption{\label{fig:range}The area in the bold black curves defines the "winning" range with $d_{min}=100$ and $cos(\alpha/2) = 0.95$. This is a win for the green player.}
\includegraphics[width=\linewidth]{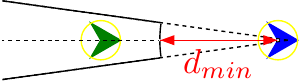}
\end{figure}

A player wins the game if it faces to the back of its opponent in a certain range before the total game ticks are used up.
Figure \ref{fig:range} illustrates how the winning range is defined.
If no one wins the game when the time is elapsed, it's a draw.

\subsubsection{Game state evaluation}
Given a game state $s$, if a ship $i$ is located in the predefined winning range (Fig. \ref{fig:range}), the player gets a score $DistScore_{i}=HIGH\_VALUE$ and the winner is set to $i$; otherwise, $DistScore_{i}=\frac{100}{dot+100}$ ($\in (0,1]$), where $dot_{i}$ is the scalar product of the vector from the ship $i$'s position to its opponent and the vector from its direction to its opponent's direction.
The position and direction of both players are taken into account to direct the trajectory. 

\ifthenelse{\longversion=1}{
\subsection{An upgraded battle game with \emph{cooldown-needless} weapon}\label{sec:game2}
We modify the previous battle game by providing a \emph{cooldown-needless} weapon and an \emph{infinity} weapon budget.
Hence, a "fire a missile" action is added to the set of legal actions.
A missile has 
\begin{itemize}
\item an identical limited remaining time $100$, which means it vanishes after $100$ game ticks;
\item a maximal speed equals to $4$ units distance ($4$ pixels in the video game) per game tick.
\end{itemize}
When launching a missile, the spaceship is affected by a recoil; when being hit by any of the opponent's missiles, the spaceship loses one life and its direction and position are changed due to an explosion, where some random process are included.
This new game is referred as $G_2$.

\subsubsection{End condition}\label{sec:end2}
The game terminates immediately if one of the following game events occurs:
\begin{itemize}
\item if only one spaceship has $life=0$ before taken the total ticks, it is a loss for the player with no lives left;
\item if both spaceships have $life=0$ before the total ticks have expired, it's a draw;
\item if both spaceships survive when the total game ticks are used up, the one with more points wins;
\item if both spaceships survive with the same number of points when the total game ticks are used up, it's a draw.
\end{itemize}

\subsubsection{Game state evaluation}
Every time a player hits its opponent, it obtains $10$ points.
Every time a player launches a missile, it is penalized by $-1$ points.
The score function is modified, considering the current number of points. 
Given a game state $s$, a player has a $nbPoints$ calculated by:
\begin{equation}
nbPoints = 10 \times nb_k - nb_m
\end{equation}
where $nb_k$ is the number of lives subtracted from the opponent and $nb_m$ indicates the number of launched missiles. The final score is a the sum of $nbPoints$ and $DistScore$ (as defined in Section~\ref{sec:game1}).

\subsection{An upgraded battle game with \emph{cooldown-required} weapon}\label{sec:game3}
Again, we modify the battle game by setting a $cooldown$ time to the weapon, i.e., after launching a missile, the weapon could not launch another during the next $cooldown$ game ticks. This version is referred as $G_3$. Both the end condition and the game state evaluation remain the same as in $G_2$.
}{}

\subsection{Perfect and incomplete information}
The battle game and its variants have \emph{perfect} information, because each agent knows all the events (e.g. position, direction, life, speed, etc. of all the objects on the map) that have previously occurred when making any decision; they have \emph{incomplete} information because neither of agents knows the type or strategies of its opponent and moves are made simultaneously.

\section{Controllers}\label{sec:cont}
In this section, we summarize the different controllers used in this work.
All controllers use the same heuristic to evaluate states (Section \ref{sec:heuristic}).

\subsection{Search Heuristic}\label{sec:heuristic}
All controllers presented in this work follow the same heuristic in the experiments where a heuristic is used (Algorithm \ref{algo:heuristic}), aiming at guiding the search and evaluating game states found during the simulations.  
\ifthenelse{\longversion=1}{
The end condition of the game is checked (detailed in Sections \ref{sec:end1} and \ref{sec:end2}) at every tick.
When a game ends, a player may have won or lost the game or there is a draw.
In the former two situations, a very high positive value or low negative value is assigned to the fitness respectively.
In the game without weapon ($G_1$), a draw only happens at the end of last tick. 
In the games with weapon ($G_2$ and $G_3$), a draw may happen earlier.
}{
The end condition of the game is checked (detailed in Section \ref{sec:end1}) at every tick.
When a game ends, a player may have won or lost the game or there is a draw.
In the former two situations, a very high positive value or low negative value is assigned to the fitness respectively.
A draw only happens at the end of last tick. 
}

\begin{algorithm}[h]
\caption{\label{algo:heuristic}Heuristic.}
\begin{algorithmic}[1]
\Function{EvaluateState}{State $s$, Player $p1$, Player $p2$}
\State{$fitness_1\gets 0$, $fitness_2\gets 0$}
\State{$s\gets Update(p1,p2)$}
\If{$Winner(s)==1$}
	\State{$fitness_1 \gets HIGH\_VALUE$}
\Else
	\State{$fitness_1 \gets LOW\_VALUE$}
\EndIf
\If{$Winner(s)==2$}
	\State{$fitness_2 \gets HIGH\_VALUE$}
\Else
	\State{$fitness_2 \gets LOW\_VALUE$}   
\EndIf
\If{$Winner(s)==null$}
	\State{$fitness_1 \gets Score_1(s)-Score_2(s)$}
	\State{$fitness_2 \gets Score_2(s)-Score_1(s)$}
\EndIf
\State{\Return{$fitness_1$, $fitness_2$}}
\EndFunction\end{algorithmic}
\end{algorithm}

\subsection{RHEA Controllers}
In the experiments described later in Section \ref{sec:xp}, two controllers implement a distinct version of rolling horizon planning: the Rolling Horizon Genetic Algorithm (RHGA) and Rolling Horizon Coevolutionary Algorithm (RHCA) controllers. These two controllers are defined next.

\subsubsection{Rolling Horizon Genetic Algorithm (RHGA)}

In the experiments described later in Section \ref{sec:xp}, this algorithm uses \emph{truncation selection} with arbitrarily chosen threshold $20$, i.e., the $20\%$ best individuals will be selected as parents. The pseudo-code of this procedure is given in Algorithm \ref{algo:bga}.
\begin{algorithm}[h]
\caption{\label{algo:bga} Rolling Horizon Genetic Algorithm (RHGA)}
\begin{algorithmic}[1]
\Require{$\lambda \in \N^*$: population size, $\lambda > 2$}
\Require{$ProbaMut \in (0,1)$: mutation probability}
\State{Randomly initialise population $\mathbf{x}$ with $\lambda$ individuals}
\State{Randomly initialise opponent's individual $y$}
\For{$x \in \mathbf{x}$}
    \State{Evaluate the fitness of $x$ and $y$}
\EndFor
\State{Sort $\mathbf{x}$ by decreasing fitness value order, so that}
$$x_1.fitness \geq x_2.fitness \geq \cdots$$
\While{time not elapsed}
    \State{Randomly generate $y$} \Comment{Update opponent's individual}
    \State{$x'_1 \gets x_1$, $x'_2 \gets x_2$}	\Comment{Keep stronger individuals}
    \For{$k \in \{3, \dots, \lambda\}$} \Comment{Offspring}
        \State{Generate $x'_k$ from $x'_1$ and $x'_2$ by uniform crossover} 
       	\State{Mutate $x'_k$ with probability $ProbMut$}
    \EndFor
    \State{$\mathbf{x}\gets\mathbf{x'}$} \Comment{Update population}
	\For{$x \in \mathbf{x}$}
    	\State{Evaluate the fitness of $x$ and $y$}
	\EndFor
	\State{Sort $\mathbf{x}$ by decreasing fitness value order, so that}
	$$x_1.fitness \geq x_2.fitness \geq \cdots$$
\EndWhile
\State{\Return{$x_1$, the best individual in $\mathbf{x}$}}\label{lst:ga}
\end{algorithmic}
\end{algorithm}

\subsubsection{Rolling Horizon Coevolutionary Algorithm (RHCA)}
RHCA (Algorithm \ref{algo:coev}) uses a tournament-based \emph{truncation selection} with threshold $20$ and two populations $\mathbf{x}$ and $\mathbf{y}$, where each individual in $\mathbf{x}$ represents some successive behaviours of current player and each individual in $\mathbf{y}$ represents some successive behaviours of its opponent.
The objective of $\mathbf{x}$ is to evolve better actions to kill the opponent, whereas the objective of $\mathbf{y}$ is to evolve stronger opponents, thus provide a worse situation to the current player.
At each generation, the best $2$ individuals in $\mathbf{x}$ (respectively $\mathbf{y}$) are preserved as parents (elites), afterwards the rest is generated using the parents by uniform crossover and mutation.
Algorithm \ref{algo:eval} is used to evaluate game state, given two populations, then sort both populations by average fitness value. Only a subset of the second population is involved.
\begin{algorithm}[h]
\caption{\label{algo:coev} Rolling Horizon Coevolutionary Algorithm (RHCA)}
\begin{algorithmic}[1]
\Require{$\lambda \in \N^*$: population size, $\lambda > 2$}
\Require{$ProbaMut \in (0,1)$: mutation probability}
\Require{$SubPopSize$: the number of selected individuals from opponent's population}
\Require{$\Call{EvaluateAndSort}$}
\State{Randomly initialise population $\mathbf{x}$ with $\lambda$ individuals}
\State{Randomly initialise opponent's population $\mathbf{y}$ with $\lambda$ individuals}
\State{$(\mathbf{x},\mathbf{y})\gets\Call{EvaluateAndSort}{\mathbf{x},\mathbf{y},SubPopSize}$}
\While{time not elapsed}
	\State{$y'_1 \gets y_1$, $y'_2 \gets y_2$}	\Comment{Keep stronger rivals}
    \For{$k \in \{3, \dots, \lambda\}$} \Comment{Opponent's offspring}
        \State{Generate $y'_k$ from $y'_1$ and $y'_2$ by uniform crossover}
       	\State{Mutate $y'_k$ with probability $ProbMut$}
    \EndFor
    \State{$\mathbf{y}\gets\mathbf{y'}$} \Comment{Update opponent's population}
    \State{$x'_1 \gets x_1$, $x'_2 \gets x_2$}	\Comment{Keep stronger individuals}
    \For{$k \in \{3, \dots, \lambda\}$} \Comment{Offspring}
        \State{Generate $x'_k$ from $x'_1$ and $x'_2$ by uniform crossover} 
       	\State{Mutate $x'_k$ with probability $ProbMut$}
    \EndFor
    \State{$\mathbf{x}\gets\mathbf{x'}$} \Comment{Update population}
    \State{$EvaluateAndSort(\mathbf{x},\mathbf{y},SubPopSize)$}
\EndWhile
\State{\Return{$x_1$, the best individual in $\mathbf{x}$}}\label{lst:coev}
\end{algorithmic}
\end{algorithm}

\begin{algorithm}[h]
\caption{\label{algo:eval}Evaluate fitness of population and subset of another population then sort the populations.}
\begin{algorithmic}
\Function{EvaluateAndSort}{population $\mathbf{x}$, population $\mathbf{y}$, $SubPopSize \in \N^*$}
\For{$x \in \mathbf{x}$}
	\For{$i \in \{1,\dots,SubPopSize\}$}
    	\State{Randomly select $y \in \mathbf{y}$}
    	\State{Evaluate the fitness of $x$ and $y$ once, update their average fitness value}
    \EndFor
\EndFor
\State{Sort $\mathbf{x}$ by decreasing average fitness value order, so that}
$$x_1.averageFitness \geq x_2.averageFitness \geq \cdots$$
\State{Sort $\mathbf{y}$ by decreasing average fitness value order, so that}
$$y_1.averageFitness \geq y_2.averageFitness  \geq \cdots$$
\Return{$(\mathbf{x},\mathbf{y})$}
\EndFunction
\end{algorithmic}
\end{algorithm}

\subsubsection{Macro-actions and single actions}
A macro-action is the repetition of the same action for $t_o$ successive time steps.
Different values of $t_o$ are used during the experiments in this work in order to show how this parameter affects the performance.
The individuals in both algorithms have genomes with length equals to the number of future actions to be optimised, i.e., $t_h$.

\def\nosue{
Detail that the operational horizon $t_o$ does not equal to the tactical/optimization horizon $t$; but $t_o$ does not necessarily equal to $1$.
$t_o>1$ may lead to two possible results:
\begin{itemize}
\item The optimization is repeated less frequently. More time can be spent on increasing the population size or the number of actions, or ...
\item This involves a bigger sub-tree saved in memory, which may be detrimental.
\end{itemize}
}

In all games, different numbers of actions per macro-action are considered in \emph{RHGA} and \emph{RHCA}. The first results show that there is an improvement in performance, for all games, the shorter the macro-action is.
The result using one action per macro-action, i.e., $t_o=1$, will be presented.

{\bf{Recommendation policy}}
In both algorithms, the recommended trajectory is the individual with highest average fitness value in the population (Algorithm \ref{algo:bga} line \ref{lst:ga}; Algorithm \ref{algo:coev} line \ref{lst:coev}).
The first action in the recommended trajectory, presented by the gene at position $1$, is the recommended action in the next single time step.

\subsection{Open Loop MCTS}
A Monte Carlo Tree Search (MCTS) using Open Loop control (OLMCTS) is included in the experimental study.  This was adapted from the OLMCTS sample controller included in the 
GVGAI distribution \cite{Perez2015}.  
The OLMCTS controller was developed for single player games,
and we adapted it for two player games by assuming a randomly acting opponent.
Better performance should be expected of a proper two-player OLMCTS version using 
a minimax (more max-N) tree policy and immediate future work is to evaluate such an agent.

{\bf{Recommendation policy}}
The recommended action in the next single time step is the one present in the most visited root child, i.e., robust child\cite{coulom2006efficient}.
If more than one child ties as the most visited, the one with the highest average fitness value is recommended.

\subsection{One-Step Lookahead}
One-Step Lookahead algorithm (Algorithm \ref{algo:osl}) is deterministic.
Given a game state $s$ at timestep $t$ and the sets of legal actions of both players $A(s)$ and $A'(s)$, One-Step Lookahead algorithm evaluates the game then outputs an action $a_{t+1}$ for the next single timestep using some recommendation policy.

\begin{algorithm}
\caption{\label{algo:osl}One-Step Lookahead algorithm.}
\begin{algorithmic}[1]
\Require{$s$: current game state}
\Require{$Score$: score function}
\Require{$\pi$: recommendation policy}
\State{Generate $A(s)$ the set of legal actions for player 1 in state $s$}
\State{Generate $A'(s)$ the set of legal actions  for player 2 in state $s$}
\For{$i \in \{1, \dots, |A(s)|\}$}
	\For{$j \in \{1, \dots, |A'(s)|\}$}
    	\State{$a_i \leftarrow i^{th}$ action in $A$}
        \State{$a_j \leftarrow j^{th}$ action in $A'$}
    	\State{$M_{i,j}\leftarrow Score_1(s,a_i,a'_j)$}
        \State{$M'_{i,j}\leftarrow Score_2(s,a_i,a'_j)$}
    \EndFor
\EndFor
\State{$\tilde a \leftarrow \pi$, using $M$ or $M'$ or both}
\State{\Return{$\tilde a$: recommended action}}
\end{algorithmic}
\end{algorithm}

{\bf{Recommendation policy}}
There are various choices of $\pi$, such as Wald \cite{wald1939} and Savage \cite{savage1951} criteria.
Wald consists in optimizing the worst case scenario, which means that we choose the best solution for the worst scenarios.
Thus, the recommended action for player 1 is 
\begin{equation}\label{eq:maxmin}
\tilde a = \argmax_{i \in \{1, \dots, |A(s)|\}} \min_{j \in \{1, \dots, |A'(s)|\}} M_{i,j}.
\end{equation}\label{eq:minmax}
Savage is an application of the Wald maximin model to the regret:
\begin{equation}
\tilde a = \argmin_{i \in \{1, \dots, |A(s)|\}} \max_{j \in \{1, \dots, |A'(s)|\}} M'_{i,j}.
\end{equation}
We also include a simple policy which chooses the action with maximal average score, i.e.,
\begin{equation}\label{eq:maxmean}
\tilde a = \argmax_{i \in \{1, \dots, |A(s)|\}} \sum_{j \in \{1, \dots, |A'(s)|\}} M_{i,j}.
\end{equation}
Respectively,
\begin{equation}\label{eq:minmean}
\tilde a = \argmin_{i \in \{1, \dots, |A(s)|\}} \sum_{j \in \{1, \dots, |A'(s)|\}} M'_{i,j}.
\end{equation}

The \emph{OneStep} controllers use separately (1) Wald (Equation \ref{eq:maxmin}) on its score; (2) maximal average score (Equation \ref{eq:maxmean}) policy; (3) Savage (Equation \ref{eq:minmax}); (4) minimal the opponent's average score (Equation \ref{eq:minmean}); (5) Wald on (its score - the opponent's score) and (6) maximal average (its score - the opponent's score).
In the experiments described in Section \ref{sec:xp} we only present the results obtained by using (6), which performs the best among the 6 policies.

\section{Experiments on battle games}\label{sec:xp}
We compare a RHGA controller, a RHCA controller, an Open Loop MCTS controller and an One-Step Lookahead controller, a \emph{move in circle} controller and a \emph{rotate-and-shoot} controller, denoted as \emph{RHGA}, \emph{RHCA}, \emph{OLMCTS}, \emph{OneStep}, \emph{ROT} and \emph{RAS} respectively, by playing a two-player battle game of perfect and incomplete information\ifthenelse{\longversion=1}{, then the same controllers are tested on some upgraded battle games with weapons.}{.}

\subsection{Parameter setting}
\begin{figure}
\centering
\includegraphics[width=.8\linewidth]{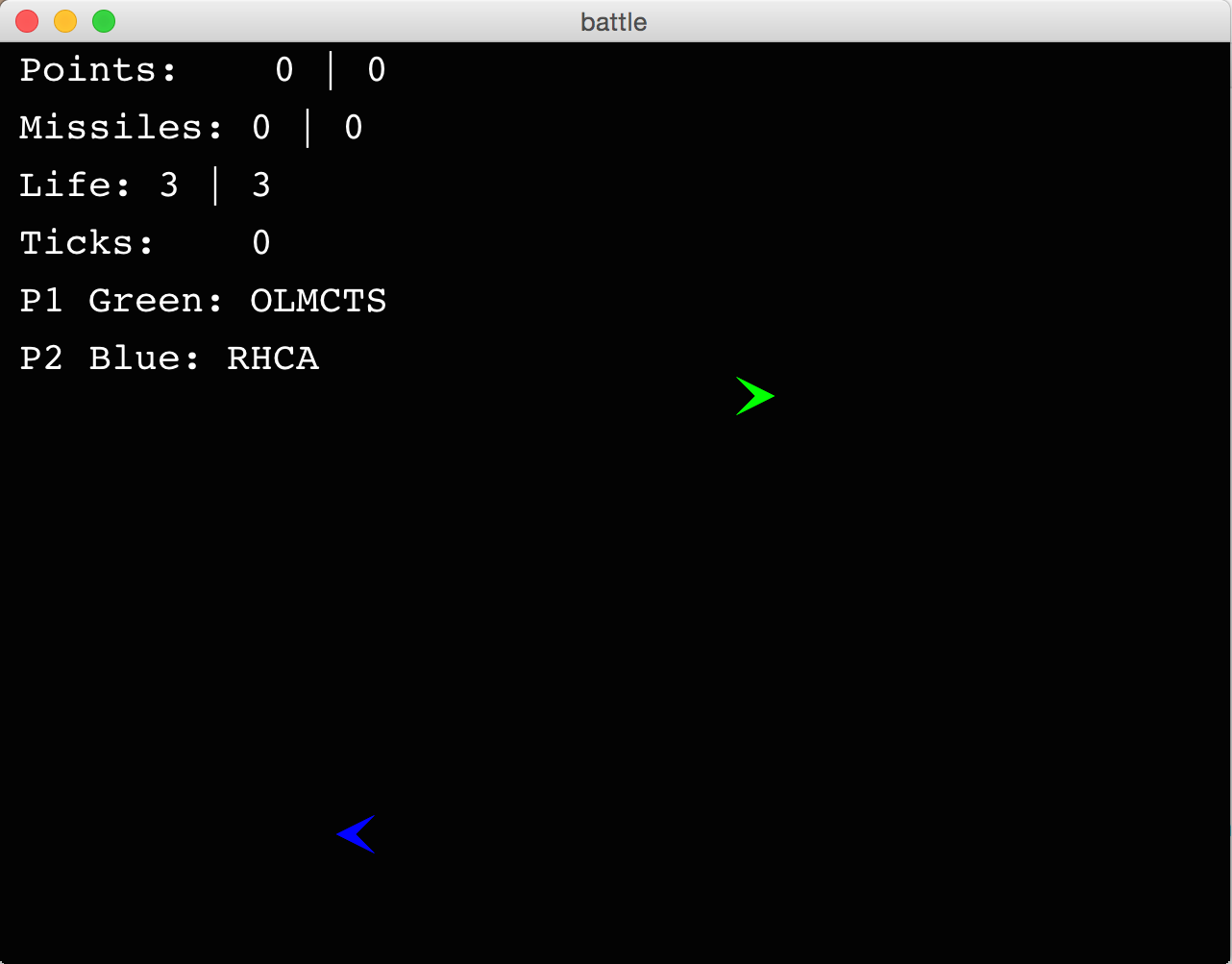}
\caption{\label{fig:initial}At the beginning of every game, each spaceship is randomly initialised with a back-to-back position.}
\end{figure}
All controllers should decide an action within $10ms$.
\ifthenelse{\longversion=1}{In games with weapon, each controller is initialized with $3$ lives.
When a $cooldown$ time is required, it is set to $4$ ticks.}
\emph{OLMCTS} uses a maximum depth$=10$.
Both \emph{RHCA} and \emph{RHGA} have a population size $\lambda=10$, $ProbaMut=0.3$ and $t_h=10$.
The size of sub-population used in tournament ($SubPopSize$ in Algorithm \ref{algo:coev}) is set to $3$.
These parameters are arbitrarily chosen.
Games are initialised with random positions of spaceships and opposite directions (cf. Figure \ref{fig:initial}).

\subsection{Analysis of numerical results}
We measure a controller by the number of wins and how quickly it achieves a win.
Controllers are compared by playing games with each other using the full round-robin league.
As the game is stochastic (due to the random initial positions of ships, the random process included in recoil or hitting), several games with random initialisations are played between any pair of the controllers.

\ifthenelse{\longversion=1}{
It's notable that the games with weapon ($G_2$ and $G_3$) are designed to kill the opponent as \emph{soon} as possible with as \emph{less missiles} as possible. 
If we only consider the number of wins, ignoring the cost of missiles, the best agent will be the one one that always goes round in circles and fires constantly (denoted as \emph{RAS}, Figure \ref{fig:ras}).
This is very important for understanding the experimental results.
\begin{figure}
\centering
\includegraphics[width=.48\linewidth]{RAS.png}
\includegraphics[width=.48\linewidth]{RAScool.png}
\caption{\label{fig:ras}\emph{RAS} controller goes in circle and shoot. Left: \emph{cooldown-needless}; right: \emph{cooldown-required}.}
\end{figure}
}

\ifthenelse{\longversion=1}{When no weapon is included in the battle game ($G_1$), a}{A} rotated controller, denoted as \emph{ROT}, is also compared.
For comparison, a random controller, denoted as \emph{RND}, is included in all the battle games.

Table \ref{tab:games} illustrates the experimental results\footnote{A video of G3 can be found at https://youtu.be/X6SnZr10yB8}.

\ifthenelse{\longversion=1}{
\begin{table}
\centering
\caption{\label{tab:games}Analysis of the number of wins of the line controller against the column controller in the battle games described in Sections \ref{sec:game1} (test case 1), \ref{sec:game2} (test case 2) and \ref{sec:game3} (test case 3). The bottom table presents the results for battle game with $cooldown$ required weapon, but the recoil is removed  (test case 4). The more wins achieved, the better controller performs. $10ms$ is given at each game tick and $MaxTick=2000$. Each game is repeated $100$ times with random initialization.
(i) $RHCA$ outperforms all the controllers in the test case $1$.
(ii) As no $cooldown$ is required, RAS dominates with a circular wall of bullets in the test case $2$.
(iii) When a $cooldown$ is required, $RHCA$ performs almost as well as RAS in the test case $3$ (the difference in terms of winning rate is $1.2\%$).
More discussion can be found in Sections \ref{sec:rot}-\ref{res:rhga}.
}
\def\rep{
\begin{tabular}{c|llllll}
\hline
\multicolumn{7}{c}{\bf{Test case 1: Simple battle game \emph{without} weapon ($G_1$)}}\\
\hline
 & RND & ROT & OneStep & OLMCTS & $RHCA$ & $RHGA$ \\
\hline
RND & - & 37.5 (784$\pm$92) & 49.0 (764$\pm$493) & 91.0 (742$\pm$53) & 93.5 (802$\pm$55) & 92.5 (775$\pm$53) \\
ROT & 62.5 (911$\pm$75) & - & 100.0 (91$\pm$3) & 100.0 (233$\pm$12) & 100.0 (160$\pm$11) & 100.0 (231$\pm$18) \\
OneStep & 51.0 (694$\pm$286) & 0.0 & - & 50.5 (1982$\pm$0) & 87.0 (780$\pm$59) & 58.0 (599$\pm$111) \\
OLMCTS & 9.0 & 0.0 & 49.5 & - & 50.0 & 50.0 \\
$RHCA$ & 6.5 & 0.0 & 13.0 & 50.0 & - & 50.0 \\
$RHGA$ & 7.5 & 0.0 & 42.0 & 50.0 & 50.0 & - \\
\hline
\hline
\multicolumn{7}{c}{\bf{Test case 2: Battle game with a $no-cooldown$ required weapon ($G_2$), considering $DistScore$ and $nbPoints$}}\\
\hline
 & RND & RAS & OneStep & OLMCTS & $RHCA$ & $RHGA$ \\
\hline
RND & - & 100.0 (323$\pm$11) & 11.0 (1940$\pm$59) & 39.0 (1105$\pm$100) & 56.0 (988$\pm$76) & 42.0 (1416$\pm$97) \\
RAS & 0.0 & - & 0.0 & 1.0 (63$\pm$0) & 4.0 (275$\pm$75) & 2.0 (124$\pm$3) \\
OneStep & 89.0 (762$\pm$50) & 100.0 (127$\pm$1) & - & 100.0 (95$\pm$4) & 100.0 (165$\pm$10) & 98.0 (200$\pm$17) \\
OLMCTS & 61.0 (862$\pm$59) & 99.0 (209$\pm$5) & 0.0 & - & 62.0 (230$\pm$19) & 50.0 (184$\pm$12) \\
$RHCA$ & 44.0 (783$\pm$66) & 96.0 (268$\pm$8) & 0.0 & 38.0 (224$\pm$19) & - & 41.0 (300$\pm$32) \\
$RHGA$ & 58.0 (896$\pm$62) & 98.0 (186$\pm$4) & 2.0 (200$\pm$101) & 50.0 (229$\pm$15) & 59.0 (342$\pm$35) & - \\
\hline
\hline
\multicolumn{7}{c}{\bf{Test case 3: Battle game with a $cooldown$ required weapon ($G_3$), considering $DistScore$ and $nbPoints$}}\\
\hline
 & RND & RAS & OneStep & OLMCTS & $RHCA$ & $RHGA$ \\
\hline
RND & - & 83.0 (921$\pm$43) & 14.0 (1916$\pm$83) & 27.0 (1355$\pm$151) & 57.0 (1130$\pm$73) & 36.0 (1700$\pm$91) \\
RAS & 17.0 (931$\pm$141) & - & 1.0 (240$\pm$0) & 30.0 (250$\pm$18) & 74.0 (513$\pm$22) & 30.0 (319$\pm$28) \\
OneStep & 86.0 (935$\pm$50) & 99.0 (295$\pm$12) & - & 100.0 (129$\pm$15) & 99.0 (178$\pm$11) & 100.0 (186$\pm$14) \\
OLMCTS & 73.0 (983$\pm$59) & 70.0 (489$\pm$17) & 0.0 & - & 60.0 (271$\pm$16) & 61.0 (232$\pm$13) \\
$RHCA$ & 43.0 (810$\pm$70) & 26.0 (578$\pm$33) & 1.0 (195$\pm$0) & 40.0 (233$\pm$13) & - & 48.0 (303$\pm$22) \\
$RHGA$ & 64.0 (1054$\pm$59) & 70.0 (421$\pm$20) & 0.0 & 39.0 (200$\pm$12) & 52.0 (302$\pm$24) & - \\
\hline
\end{tabular}
}
\scriptsize
\begin{tabular}{cccccccc}
\hline
\multicolumn{7}{c}{\bf{Test case 1: Simple battle game \emph{without} weapon ($G_1$).}}\\
\hline
 & RND & ROT & OneStep & OLMCTS & $RHCA$ & $RHGA$ & {\cellcolor[gray]{0.9}}Avg.\\
  RND  & - & 62.5 & 51.0 & 9.0 & 6.5 & 7.5 & {\cellcolor[gray]{0.9}}27.3\\
  RAS  & 37.5 & - & 0.0 & 0.0 & 0.0 & 0.0 & {\cellcolor[gray]{0.9}}7.5\\
  OneStep  & 49.0 & 100.0 & - & 49.5 & 13.0 & 42.0 & {\cellcolor[gray]{0.9}}50.7\\
  OLMCTS  & 91.0 & 100.0 & 50.5 & - & 50.0 & 50.0 & {\cellcolor[gray]{0.9}}68.3\\
  $RHCA$  & 93.5 & 100.0 & 87.0 & 50.0 & - & 50.0 & {\cellcolor[gray]{0.9}}\bf{76.1}\\
  $RHGA$  & 92.5 & 100.0 & 58.0 & 50.0 & 50.0 & -& {\cellcolor[gray]{0.9}}70.1\\
  {\cellcolor[gray]{0.9}}Avg.   & {\cellcolor[gray]{0.9}}72.7   & {\cellcolor[gray]{0.9}}92.5   & {\cellcolor[gray]{0.9}}49.3   & {\cellcolor[gray]{0.9}}31.7   & {\cellcolor[gray]{0.9}}23.9    & {\cellcolor[gray]{0.9}}29.9 & \\
\hline
\hline
\multicolumn{7}{c}{\bf{Test case 2: Battle game with a $no-cooldown$}}\\
\multicolumn{7}{c}{\bf{required weapon ($G_2$), considering $DistScore$ and $nbPoints$.}}\\
\hline
 && RND & RAS & OneStep & OLMCTS & $RHCA$ & $RHGA$ \\
 \cline{3-7}
&{\cellcolor[gray]{0.9}}Avg.&{\cellcolor[gray]{0.9}}49.6   &{\cellcolor[gray]{0.9}}1.4&{\cellcolor[gray]{0.9}}97.4&{\cellcolor[gray]{0.9}}  54.4&{\cellcolor[gray]{0.9}}43.8&{\cellcolor[gray]{0.9}}   53.4\\
RND &{\cellcolor[gray]{0.9}}50.4&-&0.0&89.0&61.0&44.0&58.0\\
RAS &{\cellcolor[gray]{0.9}}\bf{98.6}&100.0&-&100.0&99.0&96.0&98.0\\
OneStep &{\cellcolor[gray]{0.9}}2.6&11.0&0.0&-&0.0&0.0&2.0\\
OLMCTS &{\cellcolor[gray]{0.9}}45.6&39.0&1.0&100.0&-&38.0&50.0\\
$RHCA$ &{\cellcolor[gray]{0.9}}56.2&56.0&4.0&100.0&62.0&-&59.0\\
$RHGA$ &{\cellcolor[gray]{0.9}}46.6&42.0&2.0&98.0&50.0&41.0&-\\
\hline
\hline
\multicolumn{7}{c}{\bf{Test case 3: Battle game with a $cooldown$ required}}\\
\multicolumn{7}{c}{\bf{weapon ($G_3$), considering $DistScore$ and $nbPoints$.}}\\
\hline
 && RND & RAS & OneStep & OLMCTS & $RHCA$ & $RHGA$ \\
 \cline{3-8}
&{\cellcolor[gray]{0.9}}Avg. &{\cellcolor[gray]{0.9}}43.4&{\cellcolor[gray]{0.9}}30.4&{\cellcolor[gray]{0.9}}   96.8&{\cellcolor[gray]{0.9}}52.8&{\cellcolor[gray]{0.9}}   31.6&{\cellcolor[gray]{0.9}}45.0\\
RND &{\cellcolor[gray]{0.9}}56.6&-&17.0&86.0&73.0&43.0&64.0\\
RAS &{\cellcolor[gray]{0.9}}\bf{69.6}&83.0&-&99.0&70.0&26.0&70.0\\
OneStep &{\cellcolor[gray]{0.9}}3.2&14.0&1.0&-&0.0&1.0&0.0\\
OLMCTS &{\cellcolor[gray]{0.9}}47.2&27.0&30.0&100.0&-&40.0&39.0\\
$RHCA$ &{\cellcolor[gray]{0.9}}68.4&57.0&74.0&99.0&60.0&-&52.0\\
$RHGA$ &{\cellcolor[gray]{0.9}}55.0&36.0&30.0&100.0&61.0&48.0&-\\
\hline
\end{tabular}
\end{table}
}{
\begin{table}
\centering
\caption{\label{tab:games}Analysis of the number of wins of the line controller against the column controller in the battle games described in Section \ref{sec:game1}. The more wins achieved, the better controller performs. $10ms$ is given at each game tick and $MaxTick=2000$. Each game is repeated $100$ times with random initialization. $RHCA$ outperforms all the controllers.}
\scriptsize
\begin{tabular}{cccccccc}
\hline
\multicolumn{7}{c}{\bf{Test case 1: Simple battle game \emph{without} weapon ($G_1$).}}\\
\hline
 & RND & ROT & OneStep & OLMCTS & $RHCA$ & $RHGA$ & {\cellcolor[gray]{0.9}}Avg.\\
  RND  & - & 62.5 & 51.0 & 9.0 & 6.5 & 7.5 & {\cellcolor[gray]{0.9}}27.3\\
  RAS  & 37.5 & - & 0.0 & 0.0 & 0.0 & 0.0 & {\cellcolor[gray]{0.9}}7.5\\
  OneStep  & 49.0 & 100.0 & - & 49.5 & 13.0 & 42.0 & {\cellcolor[gray]{0.9}}50.7\\
  OLMCTS  & 91.0 & 100.0 & 50.5 & - & 50.0 & 50.0 & {\cellcolor[gray]{0.9}}68.3\\
  $RHCA$  & 93.5 & 100.0 & 87.0 & 50.0 & - & 50.0 & {\cellcolor[gray]{0.9}}\bf{76.1}\\
  $RHGA$  & 92.5 & 100.0 & 58.0 & 50.0 & 50.0 & -& {\cellcolor[gray]{0.9}}70.1\\
  {\cellcolor[gray]{0.9}}Avg.   & {\cellcolor[gray]{0.9}}72.7   & {\cellcolor[gray]{0.9}}92.5   & {\cellcolor[gray]{0.9}}49.3   & {\cellcolor[gray]{0.9}}31.7   & {\cellcolor[gray]{0.9}}23.9    & {\cellcolor[gray]{0.9}}29.9 & \\
\hline
\end{tabular}
\end{table}
}
Every entry in the table presents the number of wins of the column controller against the line controller among $100$ trials of battle games.
The number of wins is calculated as follows: for each trial of game, if it's a win of the column controller, the column controller accumulates $1$ point, if it's a loss, the line controller accumulates $1$ point; otherwise, it's a draw, both column and line controllers accumulates $0.5$ point.

\subsubsection{\emph{ROT}}\label{sec:rot}
The rotated controller \emph{ROT}, which goes in circle, is deterministic and vulnerable in simple battle games\ifthenelse{\longversion=1}{ without weapon ($G_1$).}{.}

\ifthenelse{\longversion=1}{
\subsubsection{\emph{RAS}}\label{sec:ras}
The \emph{RAS} controller dominates in games with \emph{cooldown-needless} weapon ($G_2$) and outperforms most of the controllers in games with \emph{cooldown-required} weapon ($G_3$).
However, it's beaten by \emph{RHCA} in $G_3$.
}

\subsubsection{\emph{RND}}\label{sec:rnd}
The \emph{RND} controller is not outstanding in the simple battle game\ifthenelse{\longversion=1}{ without weapon ($G_1$).}{.}
\ifthenelse{\longversion=1}{
It's notable that its high winning rates in the games with weapon ($G_2$ and $G_3$) is foregone.
The actions are uniform randomly chosen, without considering the cost of missiles.
As a result, it shoots more frequently than the other controllers, except \emph{RAS}.
In most of the games terminated by a win of \emph{RND}, \emph{RND} has a much lower number of points than the other controllers (except \emph{RAS}).}

\subsubsection{\emph{OneStep}}
\emph{OneStep} is feeble in all games against all the other controllers except one case: against \emph{ROT} in $G_1$.
It's no surprise that \emph{OneStep} beats \emph{ROT}, a deterministic controller, in all the $100$ trials of $G_1$.
Among the $100$ trials of $G_1$, \emph{OneStep} is beaten separately by \emph{OLMCTS} once and \emph{RND} twice, the other trials finish by a draw.
This explains the high standard error.

\subsubsection{\emph{OLMCTS}}
\emph{OLMCTS} outperforms \emph{ROT} and \emph{RND}\ifthenelse{\longversion=1}{ in games without weapon ($G_1$)}, however, there is no clear advantage or disadvantage when against \emph{OneStep}, \emph{RHCA} or \emph{RHGA}. 
\ifthenelse{\longversion=1}{
The battle game is not hard enough to distinguish the strength of controllers.
It is the main reason we upgrade the battle game by introducing weapon.}

\def\nouse{
\begin{figure}[h]
\centering
\includegraphics[width=.8\linewidth]{OLMCTS-COEV-crop.pdf}
\caption{\label{fig:olmcts}An example of \emph{OLMCTS} (green) against \emph{RHCA} (blue) in battle game with $cooldown$ required weapon ($G_3$).}
\end{figure}}

\subsubsection{\emph{RHCA}}
In all the games, the less actions set in a macro-action, the better \emph{RHCA} performs. 
\emph{RHCA} outperforms all the controllers\ifthenelse{\longversion=1}{ in $G_1$, $G_3$, and is the second-best in $G_2$, where \emph{RAS} without cooldown dominates.}{.}

\subsubsection{RHGA}\label{res:rhga}
In all the games, the less actions set in a macro-action, the better \emph{RHGA} performs.
\emph{RHGA} is the second-best in the battle game\ifthenelse{\longversion=1}{ without weapon ($G_1$); it's beaten by \emph{RHCA} when missiles are included; it performs sightly better than \emph{OLMCTS} in $G_2$ and $G_3$.}{.}

\def\nouse{
\begin{figure}[h]
\centering
\includegraphics[width=.8\linewidth]{GA-COEV-crop.pdf}
\caption{\label{fig:ga-coev}An game of \emph{RHGA} (green) plays against \emph{RHCA} (blue). Blue (respectively green) curves show the simulated trajectory of the blue (respectively green) player. White curves show the trajectory simulated by its opponent.}
\end{figure}}

\ifthenelse{\longversion=1}{
\subsection{What happens if the game is less stochastic?}\label{sec:norecoil}
In the games with weapon, the spaceship is affected by a recoil when launching a missile or an explosion when being hit by any of the opponent's missiles.
We now remove the recoil and explosion in the battle game with \emph{cooldown-required} weapon, thus make the game deterministic once the game begins, thus the initial positions of both ships are set. The new game is denoted as $G'_3$.
The same experiments are performed on $G'_3$ and experimental results are analysed in Table \ref{tab:norecoil}.
\begin{table}
\centering
\caption{\label{tab:norecoil}Analysis of the number of wins of the line controller against the column controller in the battle games without recoil or explosion (test case 4, \emph{deterministic}), as described in Section \ref{sec:norecoil}. The more wins achieved, the better controller performs. $10ms$ is given at each game tick and $MaxTick=2000$. Each game is repeated $100$ times with random initialization.
$RHCA$ performs almost as well as OLMCTS in the test case $4$ (the difference in terms of winning rate is less than $1\%$).
}
\def\rep{
\begin{tabular}{c|llllll}
\hline
\multicolumn{7}{c}{\bf{Test case 4: Battle game with a $cooldown$ required weapon ($G'_3$), without \emph{recoil}, considering $DistScore$ and $nbPoints$}}\\
\hline
 & RND & RAS & ONESTEP & OLMCTS & $RHCA$ & $RHGA$ \\
\hline
RND & - & 73.0 (763$\pm$42) & 48.0 (757$\pm$71) & 43.0 (914$\pm$99) & 69.0 (843$\pm$71) & 58.0 (1134$\pm$84) \\
RAS & 27.0 (559$\pm$91) & - & 67.5 (106$\pm$2) & 99.0 (110$\pm$1) & 97.0 (162$\pm$5) & 85.0 (124$\pm$3) \\
ONESTEP & 52.0 (743$\pm$59) & 32.5 (117$\pm$8) & - & 100.0 (63$\pm$1) & 87.0 (107$\pm$5) & 89.0 (105$\pm$5) \\
OLMCTS & 57.0 (849$\pm$71) & 1.0 (166$\pm$0) & 0.0 & - & 35.5 (113$\pm$8) & 49.5 (108$\pm$5) \\
$RHCA$ & 31.0 (700$\pm$78) & 3.0 (172$\pm$41) & 13.0 (129$\pm$16) & 64.5 (110$\pm$6) & - & 36.0 (323$\pm$40) \\
$RHGA$ & 42.0 (840$\pm$85) & 15.0 (190$\pm$13) & 11.0 (137$\pm$21) & 50.5 (102$\pm$3) & 64.0 (225$\pm$25) & - \\
\hline
\end{tabular}
}
\scriptsize
\begin{tabular}{cccccccc}
\hline
\multicolumn{7}{c}{\bf{Test case 4: Battle game with a $cooldown$ required weapon ($G'_3$),}}\\
\multicolumn{7}{c}{\bf{without \emph{recoil}, considering $DistScore$ and $nbPoints$.}}\\
\hline
 & &RND & RAS & ONESTEP & OLMCTS & $RHCA$ & $RHGA$ \\
\cline{3-7} 
&{\cellcolor[gray]{0.9}}Avg. &{\cellcolor[gray]{0.9}}58.2  &{\cellcolor[gray]{0.9}}75.1   &{\cellcolor[gray]{0.9}}72.1   &{\cellcolor[gray]{0.9}}28.6   &{\cellcolor[gray]{0.9}}29.5   &{\cellcolor[gray]{0.9}}36.5\\
RND &{\cellcolor[gray]{0.9}}41.8& -& 27.0	& 52.0	& 57.0	& 31.0	& 42.0\\
RAS &{\cellcolor[gray]{0.9}}24.9& 73.0	& - & 32.5 & 1.0	& 3.0	& 15.0\\
ONESTEP &{\cellcolor[gray]{0.9}}27.9& 48.0	& 67.5	& -		& 0.0	& 13.0	& 11.0\\
OLMCTS 	&{\cellcolor[gray]{0.9}}\bf{71.4}& 43.0	& 99.0	& 100.0	& -		& 64.5	& 50.5\\
$RHCA$ 	&{\cellcolor[gray]{0.9}}70.5& 69.0	& 97.0	& 87.0	& 35.5	& -		& 64.0\\
$RHGA$ 	&{\cellcolor[gray]{0.9}}63.5& 58.0	& 85.0	& 89.0	& 49.5	& 36.0	& -\\
\hline
\end{tabular}
\end{table}

When no random process is included, \emph{RAS} is weak against \emph{OLMCTS}, \emph{RHCA} and \emph{RHGA}.
\emph{RHCA} is beaten by \emph{OLMCTS}, however, it performs better than \emph{OLMCTS} when playing against \emph{RND} and \emph{RHGA} controller.

\subsection{What happens if the game state evaluation is modified?}\label{sec:badh}
Now we modify the way a given game state is evaluated and studied how the performances of previously used controllers vary in different games with weapon.

Instead of including the position, direction and number of points ($DistScore$ and $nbPoints$) when evaluating a given game state, we only consider $nbPoints$ in the following experiments on the games with weapon.
Table \ref{tab:gamesscore} analyses the number of wins and shows how the performance of different controllers changes when the game is evaluated differently.
However, the numerical results do not show the path of controllers, which are more or less meaningless and similar to a random controller\footnote{A video can be found at https://youtu.be/zLKO60YIKTY}.

\begin{table}
\centering
\caption{\label{tab:gamesscore}Analysis of the number of wins of the line controller against the column controllers in the battle games described in Sections \ref{sec:game2} and \ref{sec:game3}, taking into account only the number of points. The more wins achieved, the better controller performs. $10ms$ is given at each game tick and $MaxTick=2000$. Each game is repeated $100$ times with random initialization.
(i) RAS dominates in both test cases.
(ii) $RHEAs$ perform poorly due to the unsuitable heuristic, while OLMCTS achieves a winning rate around $50\%$.
}
\def\rep{
\begin{tabular}{c|llllll}
\hline
\multicolumn{7}{c}{\bf{Test case 5: Simple battle game with a $no-cooldown$ required weapon ($G_2$), considering $nbPoints$}}\\
\hline
 & RND & RAS & OneStep & OLMCTS & $RHCA$ & $RHGA$ \\
\hline
RND & - & 100.0 (304$\pm$12) & 15.0 (1844$\pm$105) & 15.0 (1798$\pm$116) & 26.0 (1331$\pm$152) & 18.0 (1532$\pm$170) \\
RAS & 0.0 & - & 33.0 (2000$\pm$0) & 0.0 & 0.0 & 0.0 \\
OneStep & 85.0 (972$\pm$49) & 67.0 (206$\pm$10) & - & 100.0 (162$\pm$33) & 26.0 (1048$\pm$112) & 10.0 (1041$\pm$213) \\
OLMCTS & 85.0 (833$\pm$53) & 100.0 (244$\pm$8) & 0.0 & - & 26.5 (1011$\pm$114) & 42.5 (1697$\pm$86) \\
$RHCA$ & 74.0 (830$\pm$52) & 100.0 (251$\pm$7) & 74.0 (1990$\pm$5) & 73.5 (1826$\pm$53) & - & 69.5 (1734$\pm$67) \\
$RHGA$ & 82.0 (936$\pm$52) & 100.0 (275$\pm$9) & 90.0 (1957$\pm$19) & 57.5 (1634$\pm$77) & 30.5 (1054$\pm$114) & - \\
\hline
\hline
\multicolumn{7}{c}{\bf{Test case 6: Battle game with a $cooldown$ required weapon ($G_3$), considering $nbPoints$}}\\
\hline
 & RND & RAS & OneStep & OLMCTS & $RHCA$ & $RHGA$ \\
\hline
RND & - & 92.0 (886$\pm$36) & 33.0 (1876$\pm$70) & 33.0 (1853$\pm$78) & 43.0 (1771$\pm$71) & 39.0 (1653$\pm$106) \\
RAS & 8.0 (1066$\pm$248) & - & 53.0 (2000$\pm$0) & 0.0 & 1.0 (1088$\pm$0) & 1.0 (2000$\pm$0) \\
OneStep & 67.0 (1188$\pm$56) & 47.0 (519$\pm$26) & - & 100.0 (144$\pm$17) & 17.0 (1346$\pm$105) & 9.0 (1214$\pm$199) \\
OLMCTS & 67.0 (1189$\pm$60) & 100.0 (641$\pm$25) & 0.0 & - & 34.5 (1265$\pm$97) & 44.5 (1768$\pm$72) \\
$RHCA$ & 57.0 (991$\pm$60) & 99.0 (770$\pm$38) & 83.0 (1984$\pm$15) & 65.5 (1823$\pm$55) & - & 67.5 (1762$\pm$64) \\
$RHGA$ & 61.0 (1015$\pm$70) & 99.0 (705$\pm$31) & 91.0 (1997$\pm$2) & 55.5 (1654$\pm$80) & 32.5 (1241$\pm$119) & - \\
\hline
\end{tabular}
}
\begin{tabular}{cccccccc}
\hline
\multicolumn{7}{c}{\bf{Test case 5: Simple battle game with a $no-cooldown$}}\\
\multicolumn{7}{c}{\bf{required weapon ($G_2$), considering $nbPoints$.}}\\
\hline
 & 		& RND & RAS & OneStep & OLMCTS & $RHCA$ & $RHGA$ \\
\cline{3-7}
&{\cellcolor[gray]{0.9}}Avg. &{\cellcolor[gray]{0.9}}34.8&{\cellcolor[gray]{0.9}}6.6&{\cellcolor[gray]{0.9}}57.6&{\cellcolor[gray]{0.9}}50.8&{\cellcolor[gray]{0.9}}78.2&{\cellcolor[gray]{0.9}}72.0\\
RND &{\cellcolor[gray]{0.9}}65.2& - 		& 0.0	& 85.0	& 85.0	& 74.0	& 82.0\\
RAS &{\cellcolor[gray]{0.9}}\bf{93.4}& 100.0 	& -		& 67.0	& 100.0	& 100.0	& 100.0\\
OneStep &{\cellcolor[gray]{0.9}}42.4& 15.0	& 33.0	& -		& 0.0	& 74.0	& 90.0\\
OLMCTS &{\cellcolor[gray]{0.9}}49.2&15.0	& 0.0	& 100.0	& -		& 73.5	& 57.5\\
$RHCA$ &{\cellcolor[gray]{0.9}}21.8&26.0	& 0.0	& 26.0	& 26.5	& -& 30.5\\
$RHGA$ &{\cellcolor[gray]{0.9}}28.0&18.0	& 0.0	& 10.0	& 42.5	& 69.5	& -\\
\hline
\hline
\multicolumn{7}{c}{\bf{Test case 6: Battle game with a $cooldown$}}\\
\multicolumn{7}{c}{\bf{required weapon ($G_3$), considering $nbPoints$.}}\\
\hline
 & &RND & RAS & OneStep & OLMCTS & $RHCA$ & $RHGA$ \\
\cline{3-7}
&\cellcolor[gray]{0.9}Avg.&\cellcolor[gray]{0.9}48.0&\cellcolor[gray]{0.9}12.6&\cellcolor[gray]{0.9}48.0&\cellcolor[gray]{0.9}49.2&\cellcolor[gray]{0.9}74.4&\cellcolor[gray]{0.9}67.8\\
RND &\cellcolor[gray]{0.9}52.0& -			& 8.0	& 67.0	& 67.0	& 57.0	& 61.0\\
RAS &\cellcolor[gray]{0.9}\bf{87.4}& 92.0		& -		& 47.0	& 100.0	& 99.0	& 99.0\\
OneStep &\cellcolor[gray]{0.9}52.0& 33.0	& 53.0	& -		& 0.0	& 83.0	& 91.0\\
OLMCTS &\cellcolor[gray]{0.9}50.8& 33.0	& 0.0	& 100.0	& -		& 65.5	& 55.5\\
$RHCA$ &\cellcolor[gray]{0.9}25.6& 43.0	& 1.0	& 17.0	& 34.5	& -		& 32.5\\
$RHGA$ &\cellcolor[gray]{0.9}32.2& 39.0	& 1.0	& 9.0	& 44.5	& 67.5	& -\\
\hline
\end{tabular}
\end{table}
}{}

\section{Conclusions and further work}\label{sec:conc}
In this work, we design a new Rolling Horizon Coevolutionary Algorithm (RHCA) for decision making in two-player real-time video games. This algorithm is compared to a number of general algorithms on the simple battle game designed, and some more difficult variants to distinguish the strength of the compared algorithms. In all the games, more actions per macro-action lead to a worse performance of both Rolling Horizon Evolutionary Algorithms (controllers denoted as \emph{RHGA} and \emph{RHCA} in the experimental study). Rolling Horizon Coevolution Algorithm (\emph{RHCA}) is found to perform the best or second-best in all the games.
\ifthenelse{\longversion=1}{
It significantly outperforms the Open Loop MCTS in battle games with weapon, though immediate future work is to test a proper two-player version of Open Loop MCTS adapted for two-player adversarial simultaneous move games.}
\ifthenelse{\longversion=1}{
Intransitivities were observed in the tournament (Section \ref{sec:norecoil}). Intransitivities have been studied in \cite{samothrakis2013coevolving}. Computing a Transitivity Index and Kullback-Leibler (KL) divergence were used to detect intransitivities.
It will be interesting to use such techniques to analyse the extent to which
transitivities are occurring in RHCA and take similar measure to cope with them.}
\def\nouse{The former provides a transitivity degree of the set of controllers; the latter is the divergence between the observed winning rate and the winning rate predicted by the Bradley-Terry (BT) model, however they studied the cycling behavior inside a population.}

More work on battle games with weapon is in progress.
In the battle game, the sum of two players' fitness value remains 0.
An interesting further work is to use a mixed strategy by computing Nash Equilibrium\cite{osborne1994course}.
Furthermore, a Several-Step Lookahead controller is used to recommend the action at next tick, taken into account actions in the next $n$ ticks.
A One-Step Lookahead controller is a special case of Several-Step Lookahead controller with $n=1$.
As the computational time increases exponentially as a function of $n$, some adversarial bandit algorithms may be included to compute an approximate Nash\cite{liu2015portfolio} and it would be better to include some infinite armed bandit technique.

Finally, it is worth emphasizing that rolling horizon evolutionary algorithms
provide an interesting alternative to MCTS that has been very much under-explored.
In this paper we have taken some steps to redress this with initial developments
of a rolling horizon coevolution algorithm.  The algorithm described here is a
first effort and while it shows significant promise, there are many obvious ways
in which it can be improved, such as biasing the roll-outs \cite{lucas2014fast}.
In fact, any of the techniques that can be used to improve MCTS rollouts
can be used to improve RHCA.  




\balance
\bibliographystyle{IEEEtran}
\bibliography{MendeleyLiu,gvgai}

\begin{thebibliography}{10}
\providecommand{\url}[1]{#1}
\csname url@samestyle\endcsname
\providecommand{\newblock}{\relax}
\providecommand{\bibinfo}[2]{#2}
\providecommand{\BIBentrySTDinterwordspacing}{\spaceskip=0pt\relax}
\providecommand{\BIBentryALTinterwordstretchfactor}{4}
\providecommand{\BIBentryALTinterwordspacing}{\spaceskip=\fontdimen2\font plus
\BIBentryALTinterwordstretchfactor\fontdimen3\font minus
  \fontdimen4\font\relax}
\providecommand{\BIBforeignlanguage}[2]{{%
\expandafter\ifx\csname l@#1\endcsname\relax
\typeout{** WARNING: IEEEtran.bst: No hyphenation pattern has been}%
\typeout{** loaded for the language `#1'. Using the pattern for}%
\typeout{** the default language instead.}%
\else
\language=\csname l@#1\endcsname
\fi
#2}}
\providecommand{\BIBdecl}{\relax}
\BIBdecl

\bibitem{coulom2006efficient}
R.~Coulom, ``{Efficient Selectivity and Backup Operators in Monte-Carlo Tree
  Search},'' in \emph{Computers and games}.\hskip 1em plus 0.5em minus
  0.4em\relax Springer, 2006, pp. 72--83.

\bibitem{kocsis2006improved}
L.~Kocsis, C.~Szepesv{\'a}ri, and J.~Willemson, ``{Improved Monte-Carlo
  Search},'' \emph{Univ. Tartu, Estonia, Tech. Rep}, vol.~1, 2006.

\bibitem{gelly2006modification}
S.~Gelly, Y.~Wang, O.~Teytaud, M.~U. Patterns, and P.~Tao, ``{Modification of
  UCT with Patterns in Monte-Carlo Go},'' 2006.

\bibitem{chaslot2006monte}
G.~Chaslot, S.~De~Jong, J.-T. Saito, and J.~Uiterwijk, ``{Monte-Carlo Tree
  Search in Production Management Problems},'' in \emph{Proceedings of the 18th
  BeNeLux Conference on Artificial Intelligence}.\hskip 1em plus 0.5em minus
  0.4em\relax Citeseer, 2006, pp. 91--98.

\bibitem{chaslot2008monte}
G.~Chaslot, S.~Bakkes, I.~Szita, and P.~Spronck, ``{Monte-Carlo Tree Search: A
  New Framework for Game AI},'' in \emph{AIIDE}, 2008.

\bibitem{Silver}
D.~Silver, A.~Huang, C.~J. Maddison, A.~Guez, L.~Sifre, G.~V.~D. Driessche,
  J.~Schrittwieser, I.~Antonoglou, V.~Panneershelvam, M.~Lanctot, S.~Dieleman,
  D.~Grewe, J.~Nham, N.~Kalchbrenner, I.~Sutskever, T.~Lillicrap, M.~Leach, and
  K.~Kavukcuoglu, ``Mastering the game of go with deep neural networks and tree
  search.''

\bibitem{perez20152014}
D.~Perez, S.~Samothrakis, J.~Togelius, T.~Schaul, S.~Lucas, A.~Cou{\"e}toux,
  J.~Lee, C.-U. Lim, and T.~Thompson, ``{The 2014 General Video Game Playing
  Competition},'' 2015.

\bibitem{mnih2015human}
V.~Mnih, K.~Kavukcuoglu, D.~Silver, A.~A. Rusu, J.~Veness, M.~G. Bellemare,
  A.~Graves, M.~Riedmiller, A.~K. Fidjeland, G.~Ostrovski \emph{et~al.},
  ``{Human-Level Control Through Deep Reinforcement Learning},'' \emph{Nature},
  vol. 518, no. 7540, pp. 529--533, 2015.

\bibitem{perez2015open}
D.~Perez, J.~Dieskau, M.~H{\"u}nermund, S.~Mostaghim, and S.~Lucas, ``{Open
  Loop Search for General Video Game Playing},'' in \emph{Proc. of the
  Conference on Genetic and Evolutionary Computation (GECCO)}, 2015.

\bibitem{weber2010optimization}
R.~Weber, ``{Optimization and control},'' \emph{University of Cambridge}, 2010.

\bibitem{ICAPS124697}
A.~Weinstein and M.~L. Littman, ``{{Bandit-Based Planning and Learning in
  Continuous-Action Markov Decision Processes}},'' in \emph{Proceedings of the
  Twenty-Second International Conference on Automated Planning and Scheduling,
  {ICAPS}, Brazil}, 2012.

\bibitem{perez2013rolling}
D.~Perez, S.~Samothrakis, S.~Lucas, and P.~Rohlfshagen, ``{Rolling Horizon
  Evolution versus Tree Search for Navigation in Single-player Real-time
  Games},'' in \emph{Proceedings of the 15th annual conference on Genetic and
  evolutionary computation}.\hskip 1em plus 0.5em minus 0.4em\relax ACM, 2013,
  pp. 351--358.

\bibitem{Perez2015}
D.~Perez, J.~Dieskau, M.~H{\"{u}}nermund, S.~Mostaghim, and S.~M. Lucas, ``Open
  loop search for general video game playing,'' \emph{Proceedings of the
  Genetic and Evolutionary Computation Conference (GECCO)}, pp. 337--344, 2015.

\bibitem{wald1939}
A.~Wald, ``{Contributions to the Theory of Statistical Estimation and Testing
  Hypotheses},'' \emph{Ann. Math. Statist.}, vol.~10, no.~4, pp. 299--326, 12
  1939.

\bibitem{savage1951}
L.~J. Savage, ``{The Theory of Statistical Decision},'' \emph{Journal of the
  American Statistical Association}, vol.~46, no. 253, pp. 55--67, 1951.

\bibitem{osborne1994course}
M.~J. Osborne and A.~Rubinstein, \emph{{A course in Game Theory}}.\hskip 1em
  plus 0.5em minus 0.4em\relax MIT press, 1994.

\bibitem{liu2015portfolio}
J.~Liu, ``{Portfolio Methods in Uncertain Contexts},'' Ph.D. dissertation,
  INRIA, 2015.

\bibitem{lucas2014fast}
S.~M. Lucas, S.~Samothrakis, and D.~Perez, ``{Fast Evolutionary Adaptation for
  Monte Carlo Tree Search},'' in \emph{Applications of Evolutionary
  Computation}.\hskip 1em plus 0.5em minus 0.4em\relax Springer, 2014, pp.
  349--360.

\end{thebibliography}
%

\end{document}